# Scalable, Energy-Efficient Optical-Neural Architecture for Multiplexed Deepfake Video Detection

Parnian Ghapandar Kashani[a]*, Shiqi Chen[a,b,c]*, Aydogan Ozcan[a,b,c†]

[a]Electrical and Computer Engineering Department, University of California, Los Angeles, CA, 90095, USA

[b]Bioengineering Department, University of California, Los Angeles, CA, 90095, USA

[c]California NanoSystems Institute (CNSI), University of California, Los Angeles, CA, 90095, USA

*These authors contributed equally to this work.

[†]Corresponding to: ozcan@ucla.edu

## Abstract

The rapid proliferation of AI-generated visual media has created an urgent need for efficient, trustworthy deepfake detection systems. However, existing deep learning–based detection methods rely on computationally intensive and energy-demanding inference algorithms, limiting their scalability. Here, we present a hybrid digital–analog deepfake video detection framework that combines a lightweight digital front-end with a spatially multiplexed optical decoding back-end for massively parallel analog inference through a programmable spatial light modulator. By simultaneously processing ≥15 video streams within a single optical-propagation pass, the system enables high-throughput and accurate video-level authenticity prediction at reduced computational cost compared with purely digital methods. We validated this hybrid deepfake video processor using different datasets spanning classical face-swapping, real-world deepfake recordings, and fully AI-generated videos. Using a spatially multiplexed experimental set-up operating in the visible spectrum, we achieved average deepfake detection accuracy, sensitivity and specificity of 97.79%, 99.86% and 95.72%, respectively, on the Celeb-DF video dataset with 15 videos tested in parallel in a single optical pass per inference. The multiplexed optical decoder also demonstrates resilience against various types of video degradation, noise, compression, experimental misalignments and black-box adversarial attacks. Our results show that integrating optical computation into AI inference enables simultaneous gains in throughput, energy efficiency, and adversarial robustness—three properties that are difficult to achieve together in purely digital systems.

## 1 Introduction

Recent advances in artificial intelligence have enabled the large-scale generation of synthetic visual content, commonly referred to as AI-generated content (AIGC) [1-6]. Modern generative models, including GAN-based face manipulation methods and more recent diffusion- and transformer-based video synthesis engines, can produce highly realistic images and videos that increasingly challenge human perception, allowing synthetic media to be created and disseminated at an unprecedented scale [7-10]. While these technologies open new possibilities for content creation, they also raise growing concerns about misinformation, identity manipulation, and the erosion of trust in digital media [11-14]. A recent study found that humans exhibited low sensitivity of ~52–55% when detecting AI-generated images, indicating a major challenge in distinguishing synthetic from real content [14-15]. This limited ability of humans to reliably identify synthetic media motivates the development of automated detection systems, which are increasingly being incorporated into media platforms and security pipelines to screen the rapidly expanding volume of visual content [16-19].

Despite significant progress made, existing deep learning-based detection systems face critical limitations in scalability and efficiency. While deep neural networks have demonstrated strong detection performance across a range of deepfake manipulation types, state-of-the-art models often require hundreds of giga floating-point operations (GFLOPs) per inference and process videos sequentially, resulting in high latency and power consumption that hinders real-time deployment at scale [20-23]. These computational bottlenecks motivate the exploration of alternative inference paradigms capable of delivering high-throughput and energy-efficient detection without relying entirely on digital processing.

Beyond these efficiency challenges, an important limitation of current deepfake detectors is their vulnerability to adversarial attacks. Extensive studies have shown that deep learning–based detection networks are inherently susceptible to targeted perturbations, such that even relatively simple manipulations of AI-generated content can cause pretrained detectors to fail entirely, rendering them unable to distinguish synthetic media from authentic content [24-26]. These challenges highlight the need for fundamentally different detection strategies. In this context, a physics-based detection paradigm that is intrinsically difficult to attack could offer a particularly compelling alternative, with the potential to provide a robust and efficient first line of defense for verifying AI-generated media.

Here, we introduce a hybrid optical–neural inference paradigm for deepfake detection that offloads key components of AI computation to physical light propagation, enabling massively parallel, energy-efficient, and intrinsically harder-to-attack inference for highly sensitive and scalable verification of AI-generated visual media. As illustrated in **Figure 1**, the proposed system combines a lightweight digital front-end for compact video representation learning with an optical decoder back-end that performs massively parallel analog inference through multiplexing on a programmable spatial light modulator (SLM). By simultaneously processing ≥15 video streams within a single optical pass through the physical system, the framework directly generates video-level authenticity predictions with reduced computational cost compared with conventional deep neural detection pipelines. This hybrid architecture enables high-throughput and highly sensitive screening of videos with low latency and provides a scalable solution for large-volume media verification. We validated this framework across different datasets spanning classical face-swapping, real-world deepfake recordings, and AI-generated videos, showing strong generalization across increasingly challenging generative AI pipelines. Using a spatially-multiplexed experimental optical set-up tested on the Celeb-DF [27] video dataset, we achieved average deepfake detection accuracy, sensitivity and specificity of 97.79%, 99.86% and 95.72%, respectively, where 15 videos in each optical pass were tested in parallel.

A key advantage of our framework is its deployment as a high-throughput screening front-end for scalable and energy-efficient deepfake detection. The proposed system is designed to operate as a computationally efficient and highly sensitive first-stage filter for rapidly detecting manipulated videos. The hybrid framework offloads part of the inference computation to a single-pass optical propagation through a compact 3D volume and exploits spatial multiplexing on an SLM to process multiple (≥15) video streams all in parallel, *i.e.*, in a snapshot. In this architecture, video processing throughput scales with the optical aperture area, offering a more favorable scaling law for large-volume media screening. Another important advantage of the proposed method lies in its resistance to physical misalignments, image noise, data compression, and adversarial attacks after deployment. Once physically implemented, the optical set-up conceals key propagation parameters and optical system configuration within the hardware, making it more difficult for attackers to infer, replicate, or manipulate the detection process. This physical barrier reduces the risk of model extraction and adversarial imitation, thereby enhancing the security and reliability of the deployed detection system. This work suggests that embedding computation within physical processes can redefine the design space of machine intelligence systems, enabling new trade-offs between performance, efficiency, and security.

# 2 Results

## 2.1 Deepfake video detection with a digital encoder and a spatially multiplexed free-space-based optical decoder

We first demonstrate that spatially multiplexed optical inference enables simultaneous processing of multiple video streams with a competitive classification performance. **Figure 1a** shows a schematic of our hybrid architecture with a lightweight digital encoder and a free-space-based optical decoder; a detailed description of the architecture is provided in the **Methods**. In our analyses, we introduce several multiplexed deepfake detection schemes that differ in one or more design parameters, including the video multiplexing factor $L$, the number of frames per video $N$, the interpolation factor $f$, and the number of diffractive layers $K$ at the optical processing back-end. For each design, we report the optimal configuration that yielded the best performance (see the **Methods** for details).

As depicted in **Fig. 1c**, cropped and standardized video frames are encoded into a 2D phase pattern in the range $[0, 2\pi)$. The encoded phase patterns of $N$ frames within a video $v$ are then spatially aggregated to form $\{\phi_{v,i}\}_{i=1}^{N}$. The digital encoder performs the same procedure for $L$ independent video streams in a batch, and the resulting combined phase seed $\Phi$ is loaded onto an SLM, collectively representing all the $L$ videos under test. For example, the top panel of **Fig. 1d** illustrates the $\Phi$ phase encoding displayed on the SLM for $N = 12$ and $L = 15$ videos. Following this batch encoding step, multiplexed optical decoding is performed under coherent illumination in a massively parallel fashion across $L = 15$ video channels simultaneously. To control cross-talk among the video channels, we adjusted the detector locations and the free-space propagation distance so that the wavefront from one video channel does not have severe cross-talk with the detectors of the adjacent video channels (see the **Methods** for details).

After the free-space propagation, the decoded output field intensity is captured by two output sensors for each video (*i.e.*, $2L$ photo-detectors in total). The bottom panel of **Fig. 1d** depicts the captured output intensity after the optical decoder, corresponding to the SLM phase pattern of the top panel. To mitigate the inherent non-negativity of light intensity, we employ a differential detection scheme [28], where the classification logit is determined by the difference between the positive and the negative detector signals at the sensor plane. The positive and negative detectors for each video channel are annotated in red and blue, respectively; see **Fig. 1d**. The mean pixel intensity value for each detector is depicted by horizontal bars. Longer red bars ($I_v^+ > I_v^-$) indicate a fake video, while longer blue bars ($I_v^- > I_v^+$) indicate a real video. We also employ a training-time temperature-scaling strategy [28,29]. This temperature hyperparameter has no effect during inference; it is used solely to accelerate training of the hybrid system (see the **Methods** for details).

Initially, we used the Celeb-DF (V2) dataset for both training and blind testing. This dataset comprises identity-based face-swap manipulations alongside the corresponding real videos [27]. After jointly training our hybrid architecture with the digital encoder and the optical decoder, we evaluated our proposed framework both numerically and experimentally. **Figure 2** illustrates the experimental set-up of our multiplexed optical deepfake detection scheme. For blind testing, we reserved 105 real and 105 fake videos from the official test split of the Celeb-DF dataset, keeping them entirely unseen during training. During testing, we repeated the frame sampling procedure for each test video using 100 different random seeds, yielding 21,000 total permutations of test videos (never seen before), which resulted in 1,400 unique SLM phase profiles with a multiplexing factor of $L = 15$ videos per inference. For each sampling instance, the batch sequence of $L = 15$ videos was randomly shuffled. This ensures that, across different frame sampling seeds, each video channel on the SLM plane is not restricted to a single type of test video but observes a diverse set of test videos, effectively covering all videos over the course of testing.

The blind testing accuracy, sensitivity, specificity, and area under the receiver operating characteristic curve (AUROC) of this design are reported in **Figure 3**, covering both channel-wise and system-level performance under numerical and experimental testing. Since each video channel operates as an independent classifier (ignoring optical cross-talk, which is minimized), the AUROC metric is reported per channel, as aggregating it across multiple classifiers is not meaningful. According to **Figure 3**, there is a strong agreement between the simulation (green) and the experimental results (blue) across all the video channels. As shown in **Figure 3**, AUROC is close to unity, and the detection accuracy remains consistently high (~97–99%) for both simulations and experimental results. Nevertheless, minor discrepancies between the simulation and experimental results can be observed. For instance, in the top panel of **Figure 3**, video channels 4 and 15 show reduced accuracy in the experimental results. Notably, these channels are located near the edges/corners of the detection field of view (see **Fig. 1d**). This suggests that the relatively increased deviations in our experimental results for these 2 video channels can likely be attributed to off-axis optical aberrations and other system-level experimental imperfections in the optical set-up. Despite these small variations, the overall experimental performance with an average sensitivity and an average specificity of 99.86% and 95.72%, respectively, demonstrates the robustness of the proposed framework under practical experimental conditions. Moreover, our fake video detection sensitivity remains near-perfect across all video channels (~99–100%) in both simulations and experimental results, indicating a nearly zero false-negative rate (~0.09% and ~0.14% on average across the numerical and experimental results, respectively). This high sensitivity is a key advantage of our system for its deployment as a first-stage, energy-efficient deepfake video screening module, as it minimizes the risk of manipulated/fake videos being missed at the initial stage of screening and ensures that suspicious samples can be reliably forwarded to subsequent, more sophisticated models for further verification and analysis; for such an application, a slightly reduced specificity of ~96% is acceptable since the burden on the system for advanced digital analysis of false positives will be minimal.

Besides these quantitative metric-based evaluations, we also provide qualitative visualization of the classification score distributions. Specifically, the histograms of the normalized differential scores $I_v$ are shown in **Figure 4**, with the decision boundary set at 0. While slight shifts are present in the experimental distributions compared to their numerical counterparts, the real and fake scores remain highly separated, as desired. To quantify this separation between the real and fake video scores, we report the Kolmogorov-Smirnov distance $D_{KS}$ [30], defined as the maximum vertical distance between the empirical cumulative distribution functions (CDFs), where $D_{KS} = 1$ indicates perfect separation. $D_{KS}$ is calculated on the experimental data to quantitatively show that although the peaks of the distribution appear closer in the experiments, the real and fake distributions remain well separated across all video channels. **Figure 4** further reveals that our experimental results achieved a mean $D_{KS}$ distance of 0.9733 with a standard deviation of 0.0042 across the $L = 15$ video channels.

To fully exploit the space-bandwidth product of our SLM, we also introduced an alternative design in which the video multiplexing factor is increased from $L = 15$ to $L = 18$, while simultaneously increasing the number of frames per video from $N = 12$ to $N = 16$; see the **Methods** for details. **Supplementary Fig. S1** shows a representative SLM phase pattern for this $L = 18$ design and its corresponding sensor-plane intensity. As in the previous design, the detector locations and the free-space optical propagation of the system are designed to minimize the optical cross-talk between adjacent video channels. The corresponding numerical simulation and experimental results for this $L = 18$ multiplexed diffractive configuration are presented in **Supplementary Fig. S2**. In this design, the AUROC metric across all the video channels remains close to unity in both simulations and experimental results. In **Supplementary Fig. S3,** we also report the channel-wise score distributions and $D_{KS}$ of the experimental results with an average of 0.9810 and a standard deviation of 0.0068.

The consistently high $D_{KS}$ values are in agreement with the similarly high AUROC scores of **Supplementary Fig. S2**, together suggesting strong separability between the two classes across all video channels. According to **Supplementary Fig. S2,** the sensitivity also remains ~100% in the experiments. However, in this $L$ =18 design, we also observe a certain degree of specificity degradation across some video channels, along with a reduced agreement between the experimental and simulation results compared to the 15-video multiplexed design. This degradation can be attributed to a relatively increased optical cross-talk arising from the more compact packing of 18 video channels at the SLM aperture. This effect is also evident from the reduced uniformity of channel-wise performance, with an average accuracy of 96.46% and a standard deviation of 0.52% in simulations and an average accuracy of 96.13% and a standard deviation of 1.33% in our $L$ =18 experimental results, compared to an average accuracy of 97.79% and a standard deviation of 0.62% in our $L$ =15 experimental results. Overall, these results indicate that the $L$ =15 design provides a more balanced trade-off between sensitivity and specificity, and presents more uniform multiplexed performance across different video channels.

## 2.2 Energy efficiency and performance comparison of the hybrid digital-optical system

To further quantify the performance of our spatially multiplexed optical decoder, we compared it against two lightweight digital decoding strategies optimized for deepfake video classification. For the first digital alternative, we retained the same encoder in our architecture and replaced the optical decoder with a tree-based classifier, implemented via feature engineering [31]. For the second digital alternative, we again retained our digital encoder, but introduced a deep digital decoder consisting of convolutional layers, fully connected layers, and nonlinear activation functions. Detailed performance and power consumption comparisons of these two fully digital architectures are provided in **Supplementary Fig. S4**. All the reported performance and power consumption values are calculated per video with $N = 12$ frames. In this analysis, we designed the decoders to be as lightweight as possible to enable a fair comparison in the low-energy regime, while aiming for enhanced performance. More complex digital architectures, such as transformer-based digital decoders, could also be considered to improve the cross-dataset generalization; however, they are significantly more energy-intensive and fall outside the scope of this comparison. According to **Supplementary Fig. S4a**, under the same shared encoder architecture, the optical decoder consistently maintains high accuracy, sensitivity, and specificity, remaining competitive with respective to the performance of the deep digital decoder; furthermore, it significantly outperforms the tree-based digital decoder. As shown in **Supplementary Fig. S4b**, the spatially multiplexed optical decoder with $L = 15$ lies on the Pareto-optimal frontier of the accuracy-energy trade-off curve, achieving a high experimental (*numerical*) accuracy of 97.79% (*98.12%*) and a low energy consumption of 1.38-4.11 mJ per video. For the $L = 18$ design, the overall energy consumption of the optical decoder further reduces to ~1.15-3.43 mJ per video with an average experimental (*numerical*) accuracy of 96.13% (*96.46%*). In comparison, the tree-based classifier operates at a much lower energy (~0.007 mJ) but with significantly reduced accuracy (62.62%). The deep digital decoder, while achieving a high accuracy (98.24%), does so at a substantially higher energy cost (~37.36 mJ per video).

Similar to the decoder analysis, we also evaluate the performance–energy trade-off for the digital encoder of varying sizes, under a shared optical decoder with $K = 0$ and $L$ = 15 spatially multiplexed videos. We modified the digital encoder by removing some of the convolutional layers from the spatial CNN and Fourier-domain CNN modules (see the **Methods**), reducing its size and energy footprint while keeping the optical decoder architecture fixed. **Supplementary Fig. S5** reports accuracy, sensitivity, and specificity at various energy consumption levels per video. As the encoder energy consumption per video is decreased by more than 2-fold, from 178.20 mJ to 85.27 mJ, the accuracy and the specificity decrease by ~5.49% and ~10.83%, respectively, whereas the sensitivity drops by only ~0.08%. Our analysis in **Supplementary Fig. S5** suggests the trade-off is predominantly between energy consumption and specificity, with sensitivity remaining largely preserved even at significantly reduced energy budgets at the digital encoder; for example, sensitivity can be maintained at 97.45% at 39.37 mJ encoder energy consumption per video. The preservation of very high sensitivity at lighter digital encoder configurations ensures that our hybrid framework can still serve as an effective first-stage screening module without missing fake videos, where a moderate reduction in specificity is acceptable since false positives can be forwarded to a more advanced downstream model for further verification/analysis.

### 2.3 Robustness analysis under video degradation, noise, compression and physical misalignments

We conducted ablation studies to evaluate the resilience of the proposed framework under real-world deployment imperfections. First, we applied different types of perturbations (never seen during training) to input test video frames, including random Gaussian noise, Gaussian blur, and JPEG compression. Such perturbations and noise terms are particularly important to be evaluated since widely distributed or shared fake videos may inevitably undergo degradation with blurring, compression artifacts or random noise during transmission across different platforms. State-of-the-art digital models often exhibit limited robustness to such frame-level degradations that were never seen during training [32]. **Figure 5** reports the detection performance of our hybrid processor across low, moderate, and strong levels of perturbation under each category. Fake video detection accuracy, sensitivity, and specificity are plotted as a function of the standard deviation $\sigma$ of Gaussian noise (**Fig. 5a**) and Gaussian blur (**Fig. 5b)**, and the JPEG quality factor $Q$ (**Fig. 5c**) as defined by the JPEG standard [33]. As shown in **Fig. 5c,** all three performance metrics remain above ~93% across most JPEG compression levels, dropping only noticeably for $Q$<30, where compression artifacts become clearly visible. For Gaussian blur, the hybrid model maintains an accuracy above ~94% and ~78% in the mild and moderate regimes, respectively, and slowly degrades to ~60% in the strong-blur regime. The Gaussian noise presents the most impactful degradation, particularly for the detection sensitivity, which remains above ~95% for $\sigma = 0 - 0.05$, but drops sharply to below ~50% under stronger noise levels. Overall, the hybrid model remains competitive across different types of image degradation not encountered during training. Importantly, most real-world artifacts fall within the mild and moderate regimes, where the hybrid system maintains stable performance across all metrics. These results suggest that the proposed framework is robust to common forms of video degradation and can operate reliably under practical deployment conditions.

Next, we conducted a physical ablation study to evaluate the performance of the hybrid framework against 3D misalignments of the SLM and the output sensor plane. This ablation is also important for real-world deployment, such as in data centers, where thermal expansion or mechanical mounting errors may introduce positional deviations over time. In general, diffractive optical processors can be made resilient to misalignments through a vaccination training strategy, where lateral or axial misalignments are randomly introduced during the learning/training phase [34]. In **Supplementary Fig. S6**, we report the performance of both the baseline and vaccinated diffractive models under random lateral and axial misalignments, demonstrating the resilience of the optical decoder architecture. As shown in **Supplementary Fig. S6,** the vaccinated model maintains near-ideal accuracy, sensitivity, and specificity under lateral misalignments within a range of (-$200$, $200$) $\mu m$. The baseline model, on the other hand, exhibits performance degradation beyond a misalignment range of (-96, 96) $\mu m$. Furthermore, the system remains resilient under axial misalignments on the order of 0–2000 $\mu m$, with only minor performance variation of less than ~0.05% across both the baseline and the vaccinated models, confirming robustness under realistic deployment conditions. These analyses indicate that incorporating vaccination during system training further enhances the robustness of the optical decoder under various deployment conditions.

**2.4 Testing on deepfake videos generated by text-to-video models**

While existing state-of-the-art deepfake detection methods have demonstrated strong performance and cross-dataset generalization on conventional benchmarks such as FF++ [16], DFDC [17], and Celeb-DF [27], these datasets predominantly consist of manipulated real videos, particularly face-swap content, which often contain residual spatial and temporal artifacts from the digital manipulation pipeline. Although more challenging datasets, such as Celeb-DF that we used in our analyses (reported above) attempt to mitigate these artifacts through higher resolution, refined blending, and temporal smoothing, such artifacts remain fundamentally tied to the underlying manipulation pipeline [27]. In contrast, recent advances in text-to-video (T2V) generation have significantly transformed synthetic media creation, with modern models producing videos that approach photorealistic quality, largely eliminating the artifacts present in earlier methods [35]. However, large-scale benchmarks and robust detection methods for this emerging class of generative media remain limited partially due to their rapid evolution, leaving the effectiveness of existing deepfake detection approaches on this new class of generative media largely unexplored [36].

To test our spatially multiplexed optical decoder architecture with T2V generative models, we created our custom dataset of fake videos generated by the Google VEO-3 models using the publicly available Gemini interface [7]. The creation of this dataset is detailed in the **Methods** section. In this analysis, we demonstrate that lightweight fine-tuning of our hybrid processor for only <1% of the full training pass of the original model, with only 50 Gemini videos, is sufficient to achieve competitive fake video detection performance on VEO 3-generated new content, showing that the framework can be readily adapted to emerging generative synthesis pipelines. Similar to Celeb-DF blind testing reported in the earlier sub-sections, here we set aside 105 real and 105 fake videos created through Gemini (never used in training or transfer learning), with 100 sampling seeds, resulting in a total of 21,000 test video permutations with 1400 distinct SLM phase patterns using $L = 15$ video multiplexing per snapshot inference. The accuracy, sensitivity, specificity, and AUROC metrics of our blind testing on this Gemini-based fake video dataset are reported in **Figure 6**. As shown in the top panel of **Fig. 6**, the average accuracy in the numerical simulations (green dashed line) reaches 95.32% with a standard deviation of 0.63%, while the experimental results with $L = 15$ video multiplexing (blue dashed line) closely follow with an average accuracy of 94.80% and a standard deviation of 0.78%. Similar to the results on the Celeb-DF dataset, minor discrepancies are observed between our simulations and experimental results, with the most pronounced deviation occurring in the video channel 12, located near the bottom edge of the detection field of view, likely due to off-axis aberrations and experimental non-idealities in our set-up. Compared to Celeb-DF, a small reduction in accuracy, sensitivity, and specificity across all channels is observed, as the model underwent minimal fine-tuning on this new video dataset generated through Gemini and the underlying video synthesis method is fundamentally different from that of Celeb-DF (see **Methods** for details). Despite these changes, our experimental system with $L = 15$ video multiplexing demonstrates robust performance on the Gemini-based fake video dataset with an average experimental sensitivity and specificity of 97.61% and 92.00%, respectively. **Figure 7** further illustrates the normalized differential scores across different video channels for real and fake videos in the test set. Same as before, Kolmogorov-Smirnov distance $D_{KS}$ is reported per channel to quantify the separation of real and fake videos. **Figure 7** reveals that our experimental results achieved a mean $D_{KS}$ distance of 0.9087 with a standard deviation of 0.0138 across the $L = 15$ video channels for the VEO 3-generated dataset.

Furthermore, we also fine-tuned and evaluated the alternative $L = 18$ video multiplexing design on the same Gemini-generated video dataset, both numerically and experimentally. The corresponding results of this transfer learned $L = 18$ design are reported in **Supplementary Figs. S7 and S8.** Our experimental results in this $L = 18$ configuration show comparable fake video detection accuracy with respect to the $L = 15$ design that is transfer learned: 95.16% vs. 94.80% on the Gemini-generated video dataset, respectively. In **Supplementary Fig. S8,** we also report the channel-wise score distributions as well as the CDFs in the inset. The vertical black bar indicates an experimental $D_{KS}$ with a mean of 0.9437 and a standard deviation of 0.0094.

These results are particularly significant given that VEO-3 (Gemini) generated videos lack the traditional spatial and temporal artifacts typical of earlier deepfake generation pipelines. Our hybrid processor's ability to achieve high accuracy with minimal fine-tuning (<1% of a training pass) indicates that the optical decoder is extracting fundamental physical features rather than simply over-fitting to specific digital manipulation signatures.

### 2.5 Enhancing the performance of spatially multiplexed optical decoders using optimized structured diffractive layers

Next, we explored enhancing the deepfake video detection performance of the hybrid set-up using optimized structured diffractive layers within the optical decoder architecture. For this analysis, we selected a more challenging video dataset, DeepSpeak [37] as our test bed. This dataset is more challenging because a Contrastive Language–Image Pretraining (CLIP)-based identity matching is performed prior to face-swap. In the identity-matching phase, only perceptually similar identities, based on both facial and vocal features, are paired and swapped, resulting in more compelling, harder-to-detect deepfakes than Celeb-DF [37]. Thus far, the presented results have utilized free-space propagation as the spatially multiplexed optical decoder, *i.e.*, $K = 0$ was used. This hybrid model with a free-space based decoder ($K = 0$) showed reduced performance on this new dataset, with an average video detection accuracy ~77%. However, the complexity of the optical decoder can be increased without a penalty on the energy consumption or inference latency to enhance its detection performance while keeping the digital encoder unchanged. We show that the classification performance on this dataset, DeepSpeak, can be improved by increasing the decoder depth, or equivalently, by introducing $K \geq 1$ diffractive layers. **Figure 8a** presents the accuracy, sensitivity, and specificity of our hybrid architecture as a function of $K$, blindly evaluated on the DeepSpeak test dataset. These analyses reveal that the deepfake video detection accuracy increases by ~10% from $K = 0$ to $K = 1$ and further increases to ~88% at $K = 2$. In **Figure 8b**, we also report the real and fake score distributions for $K = 0, 1,$ and $2$, with the Kolmogorov–Smirnov distances $D_{KS}$ also shown in the insets. At $K = 0$, $D_{KS} = 0.737$, which is lower than the values observed for Celeb-DF and Gemini test videos, reflecting the more challenging nature of this dataset. Nevertheless, as the number of structurally optimized diffractive layers in the optical decoder increases, the statistical separation improves progressively, reaching $D_{KS} = 0.761$ at $K = 1$ and $D_{KS} = 0.793$ at $K = 2$. This trend can be visually verified in **Figure 8b**, where the real and fake distribution peaks become more clearly separated as $K$ increases.

Overall, these results demonstrate that increasing the number of passive diffractive layers leads to better statistical separation, higher classification confidence, and improved deepfake detection accuracy on the DeepSpeak dataset. It is important to note that these diffractive layers can be implemented as passive, static/frozen structures rather than reconfigurable SLMs (**Fig. 1b**) since they can be fabricated using optical lithography tools and remain fixed during inference; therefore, the total energy footprint and latency of the optical decoder remain largely unaffected compared to the vanilla version with $K = 0$. Also see **Supplementary Fig. S9** for a detailed comparative analysis of power consumption in these architectures. This comparison specifically focuses on the decoder architecture, excluding the encoding phase SLM, as the digital encoder is common and remains unchanged in each case. In the digital twin, the optical decoder incurs a relatively high number of FLOPs due to computationally intensive 2D FFT operations and elementwise multiplication of the complex-valued wave functions with the diffractive layers; see the **Methods** section. This cost grows rapidly for $K = 2$, since adding more layers requires more consecutive FFTs for free-space propagation between layers, resulting in rapid growth of power consumption in the digital twin compared to the optical *in situ* model. The power consumption of the optically implemented physical model, for the decoder part, remains constant for $K = 1, 2$ because the diffractive layers are passive static structures, which can be fabricated using different optical lithography approaches [38-41].

### 2.6 Adversarial attacks on spatially multiplexed optical decoders

State-of-the-art deepfake detectors remain vulnerable to adversarial attacks, where small and imperceptible engineered noise patterns added to the input video frames can deceive the deepfake detectors into misclassification, effectively evading the detection entirely [42]. To evaluate the robustness of the presented spatially multiplexed optical decoders to such digitally engineered attacks, we conducted an ablation study using universal [43] and transferable black-box adversarial attacks [44] across optical decoders of varying complexity ($K = 0, 1, 2$) as well as a fully digital counterpart with a deep digital decoder. Specifically, we performed black-box attacks applied by 10 independent attackers and report the performance across different perturbation budgets denoted as $\varepsilon$, which imposes a pixel-wise constraint on the strength of the applied perturbations (see the **Methods** section for details). In each attack, we restricted this perturbation budget to $\varepsilon \in \{1, 2, 4, 8\}/255$ to ensure that the attacks remain almost imperceptible to the human eye (see **Supplementary Fig. S10)**. In this black-box setting, we utilized an Xception-based [45] detection model as the surrogate network to optimize the attacks via Projected Gradient Descent (PGD) [44] (see the **Methods**). After optimizing the global frame-level perturbations under different $\varepsilon$ values, we applied them on our test video set, processed by both the hybrid digital-optical models with different $K$ values and the fully digital alternative model with a deep digital decoder. For each attack, we evaluated the accuracy, sensitivity, and specificity metrics as a function of the attack strength, $\varepsilon$. This comparative analysis, along with the visualization of the adversarial perturbations, is reported in **Supplementary Fig. S10**. These analyses reveal that as we increase the complexity of the optical decoder with increasing $K$, the hybrid model exhibits enhanced robustness to unseen adversarial perturbations compared to its fully digital counterpart. For example, **Supplementary Fig. S10** shows that the hybrid model implemented with structurally optimized diffractive layers achieves 10-25% higher specificity and 5-12% lower attack success rate (higher accuracy), compared to the fully digital model under attack. As expected, the overall accuracy of deepfake video detection decreases at higher $\varepsilon$ values as the attacks become stronger. Nevertheless, even at the highest $\varepsilon$ value, both the specificity and accuracy of our hybrid system with the diffractive decoder $K = 2$ remain superior to those of the $K = 0$ and deep digital decoder, highlighting the inherent robustness of the optical decoder architecture with optimized diffractive layers to black-box adversarial attacks.

This enhanced resilience stems from the physical concealment of the model parameters. In a $K \geq 1$ architecture, an attacker faces a significant metrology challenge: measuring the fixed decoder phase topology at the diffraction limit of light is required to launch a white-box attack. This elevates the cost of adversarial manipulation from a purely algorithmic task to one that requires precise physical characterization, which is significantly more difficult in practical settings.

## 3 Discussions

The rapid growth of AI-generated visual media has created an urgent need for efficient and reliable deepfake detection systems. Current deep learning approaches, however, depend on computationally intensive and energy-demanding inference, limiting their scalability. In this work, we demonstrated a hybrid digital–analog deepfake video detection framework that combines a lightweight digital front-end with a spatially multiplexed optical decoding back-end, enabling massively parallel analog inference using an SLM. By processing at least 15 video streams within a single optical propagation pass, this spatially multiplexed system delivers high-throughput, accurate video-level authenticity predictions while reducing computational cost relative to fully digital methods. We validated the framework on three datasets encompassing classical face-swapping, real-world deepfake recordings, and fully AI-generated videos using Gemini. In a visible-spectrum experimental set-up with $L = 15$ spatial multiplexing of test videos (Celeb-DF), the system experimentally achieved an average deepfake detection accuracy, sensitivity and specificity of 97.79%, 99.86% and 95.72%, respectively; the experimental accuracy under $L = 18$ spatial multiplexing of the same test videos slightly reduced to 96.13% due to, *e.g.*, increased optical cross-talk among video channels. While demonstrated for deepfake detection, this paradigm is applicable to a wider class of high-throughput inference tasks, including real-time surveillance, large-scale content moderation, and other security-critical AI systems.

The presented spatially multiplexed optical deepfake detection framework is designed to serve as an energy-efficient and highly sensitive first-stage screening module for large-scale deepfake detection. Therefore, a key question is how much energy can be saved compared with conventional digital models. For this comparison, we used the same Celeb-DF (V2) dataset [27] to train both a tree-based baseline model and a compact GPU-based baseline model for comparison, where the optical decoder is replaced with a digital decoder that is deep enough to match the inference performance of the hybrid model. As shown in **Supplementary Fig. S4,** when the GPU-based deep digital decoder baseline achieves detection performance comparable to that of the proposed method, our hybrid framework with the optical decoder demonstrates important energy advantages for each batch inference. In contrast, although the tree-based baseline consumes less energy, this comes at the cost of a significantly reduced detection accuracy. In **Supplementary Fig. S9**, we further compare the proposed optical decoder-based deepfake detection framework with its digital twin implementation where the end-to-end forward model of our hybrid approach is executed in silico. In this comparison against its digital twin, the physical system achieves lower energy consumption per batch than its digital counterpart, and as the complexity of the optical decoder increases (with larger $K$ to tackle more challenging datasets), this energy advantage becomes increasingly more pronounced. As for the digital encoder energy consumption, our analyses in **Supplementary Fig. S5** indicate that the detection sensitivity remains robust even at significantly reduced digital encoder energy budgets; for example, a sensitivity of ~97.45% can be achieved for an encoder energy consumption of <40 mJ per video. By maintaining high sensitivity during low-power digital encoder configurations, our hybrid framework functions as an efficient first-stage screening module that minimizes the risk of missed detections. In this context, a moderate reduction in specificity due to a more compact digital encoder is an acceptable compromise, as any false positives can be filtered by a more sophisticated downstream digital model for final verification.

In addition to energy consumption, another critical challenge in deepfake detection is the vulnerability of final predictions to image/video noise, compression and adversarial attacks or malicious tampering. For conventional digital models, once the network weights are compromised, the output can be intentionally manipulated. In contrast, for the spatially multiplexed optical decoder, the physical parameters of the system are inherently difficult to measure with sufficiently high precision even if the hardware is compromised and is in the hands of an attacker; this becomes nearly impossible with $K \geq 1$ optical decoder architectures since measurement of the fixed decoder phase topology at diffraction limit of light is a major metrology challenge making reverse engineering substantially more challenging and consequently rendering the system less susceptible to white-box attacks if the optical set-up is compromised. This positions physical computation not only as an efficiency tool, but as a foundational component for designing trustworthy AI systems.

All in all, this work demonstrates an energy-efficient optical processor that serves as a high-throughput screening front-end with favorable scaling characteristics, enabling multiplexed verification of ≥15 video streams within a single optical propagation step while offering enhanced resistance to reverse engineering or adversarial attacks after deployment. We validated the framework across different datasets, showing strong generalization across increasingly challenging synthesis pipelines, also highlighting the role of optical decoder complexity in improving deepfake detection performance. These characteristics make the proposed framework particularly well-suited as a first-stage screening front-end for scalable deepfake detection systems. More broadly, this work highlights the potential of hybrid optical–electronic computing as a scalable and physically secure paradigm for trustworthy AI systems. We envision that such architectures may serve as an enabling foundation for future real-time, large-scale verification of AI-generated media and other security-critical inference tasks.

# 4 Methods

### Deepfake and AI-generated video datasets

A typical deepfake challenge is the face-swap identity-based deepfakes, a manipulation scenario in which the identity of a target face in a video is replaced with that of another person (the donor or source), while preserving the target's facial expressions, pose, and head motion. Face-swap typically relies on autoencoder-based architectures or GANs to transfer identity features from the donor onto the facial geometry and motion of the target video [27].

To train and evaluate our framework, we considered two large-scale deepfake detection datasets comprising face-swap manipulations. First, we employed Celeb-DF (v2) [27] to train our multiplexed deepfake video detection model with a free-space-based decoder ($K = 0$).

Celeb-DF is a deepfake benchmark consisting of high-quality face-swap videos generated using an improved autoencoder-based synthesis pipeline, incorporating techniques such as color-consistency correction, context-aware blending, and temporal smoothing. In addition to Celeb-DF, we extended our framework to the DeepSpeak dataset [37], a more recent benchmark designed to reflect newer deepfake generation pipelines and more diverse video recording environments where subjects perform more realistic and less constrained gestures during the recording. DeepSpeak also performs a CLIP-based identity matching technique prior to face-swap and uses multiple new video and audio synthesis engines to generate high-quality deepfakes.

In addition to these face-swapping datasets, we constructed a custom video dataset using Google's VEO-3 text-to-video model, accessed through Gemini [7]. Unlike traditional face-swap datasets, where synthetic content is generated by transferring identity onto an existing video, VEO-3 produces videos directly from text prompts without conditioning on real video inputs, which is another typical situation that deepfake has been widely applied. To construct this dataset, we first collected *real* videos from Celeb-DF and processed them using a large language model (GPT-5) with a structured prompt to extract detailed semantic descriptions of each video. These descriptions capture attributes such as facial appearance, pose, lighting conditions, scene context, and temporal dynamics. The resulting prompts were then used to generate *synthetic* videos using the VEO-3 model. These prompts do not reference named or identifiable individuals. Thus, this dataset enabled us to evaluate our framework on some of the most recent text-to-video deepfakes that fundamentally differ from traditional identity-based face-swap manipulations.

**Experimental set-up**

The optical deepfake detection system with a free-space-based decoder with $L = 15$ design was experimentally validated using the set-up shown in **Fig. 2**. Specifically, **Fig. 2a** presents a simplified 3D rendering of the optical layout, and **Fig. 2b** shows a photograph of the experimental set-up. A laser was used as the illumination source at a wavelength of $520\,nm$. The laser beam first passed through a cylindrical lens for beam shaping, where the focal length along one axis was $100\,mm$, and then through a linear polarizer to align the polarization direction with the operating axis of the liquid-crystal array in the SLM. After polarization control, the beam entered an optical subsystem for spatial filtering, beam expansion, and collimation. This subsystem consisted of a $30\times$ microscope objective, a $100\,\mu m$ pinhole placed at the Fourier plane, and a plano-convex lens with a focal length of $200\,mm$. After passing through this subsystem, the beam was expanded and collimated to provide uniform illumination for the subsequent phase modulation stage. The processed beam was then incident on an SLM (HOLOEYE PLUTO-2.1, pixel pitch $8\,\mu m$, resolution $1920 \times 1080$), onto which the phase pattern corresponding to the deepfake video phase $\Phi(x, y)$ was loaded. Each phase structure combined the information from 15 independent videos. The SLM was operated at a refresh rate of $60\,Hz$.

The phase-modulated optical field was first relayed through a 4f system composed of two plano-convex lenses, each with a focal length of $100\,mm$. To better match the overall distribution of the $L$ detection regions to the effective sensor area of the camera, an additional lens with a focal length of $125\ mm$ was introduced after the conjugate plane to slightly rescale the image. The resulting optical field was then recorded by a camera (Basler ace acA1920-40 $\mu m$ , pixel pitch $5.86\ \mu m$, resolution $1936 \times 1216$). The effective resolution of the encoded phase pattern was $1920 \times 1056$, which was zero-padded to match the full SLM resolution. The camera exposure time was set to $0.01s$. Because the spatial dimension of the camera sensor was slightly smaller than the total size of the 15 detection regions, demagnification (using a lens with $125\,mm$ focal length) is implemented to capture all detection results simultaneously. During data collection, we first captured one frame when the SLM was flat. This capture was then blurred and divided by the mean intensity to calculate the per-pixel illumination adjustment, which accounts for potential illumination nonuniformities or static aberrations in the optical set-up. Each captured image was first corrected using the illumination adjustment, then cropped according to the pre-assigned locations of the video detector regions. For each detector region, the average intensity was calculated. The final classification result was then determined by comparing the average intensities of the corresponding positive/negative detector regions.

**Energy consumption of a spatially multiplexed optical decoder**

The energy consumption of the digital encoder (which was commonly employed in both the hybrid approach and in the other digital counterparts compared in this work) is estimated using a power efficiency of 5.5 pJ/FLOP for 32-bit floating-point operations on an NVIDIA GeForce RTX 4090 [50]. Based on this, the lightweight digital encoder consumes ~2.7 GFLOPs per frame, corresponding to 178.2 mJ per video for $N = 12$ input frames; also see **Supplementary Fig. S5** for lower-energy-consuming digital encoder architectures with a modest compromise in accuracy while retaining the sensitivity performance (e.g., achieving 97.45% sensitivity at 39.37 mJ per video). For the optical decoder, we explore the typical power budget of various off-the-shelf components that can be used to set up a baseline optical hardware for the hybrid system architecture. Because the frame rate is mostly limited by the refresh speed of the SLM, without loss of generality, here we assume a modest frame rate of 120-180Hz [51,52]. For the SLM, the system-level power consumption is conservatively estimated to be on the order of 3.6-7W [53] (including the display digital drive and the power dissipation of the LCOS panel). For the illumination source, a typical laser diode would consume <100mW [53,54]. For the signal detection, we can use $L \times 2$ photo-detectors, with a total power consumption of 30-300mW range for $L = 15$ [55]. Therefore, the overall energy consumption of a baseline spatially multiplexed optical decoder using off-the-shelf optical components would be 3.73-7.40W, resulting in 20.7-61.7mJ per detection batch and 1.38-4.11 mJ per video ($L = 15$). For the $L = 18$ design, the overall energy consumption of a spatially multiplexed optical decoder would further reduce to ~1.15-3.43 mJ per video. The numbers are conservatively estimated based on commercially available optical components, and can be further reduced with optimized, state-of-the-art optical hardware.

**Formulation of black-box attacks**

To formulate our attack, we simulate 10 independent attackers. Each attacker $m \in \{$ 1, 2,

…,10} is assigned a distinct random seed and granted access to a class-balanced subset $D^{(m)}$ of Celeb-DF training data. The attacker has no access to the digital encoder weights or the optical decoder forward model. Instead, each attacker uses XceptionNet [45] as a surrogate model, denoted as $f_{surr}(.)$ to learn the universal attack pattern. In this formulation, the attacker $m$ learns a universal $\delta^{(m)} \in \mathbb{R}^{3 \times H \times W}$, which is a spatial pattern stamped identically on every frame of a video with lateral dimensions of $H \times W$. The perturbation is constrained to be almost imperceptible by bounding its maximum per-pixel magnitude: $\|\delta^{(m)}\|_\infty \leq \varepsilon$, where $\varepsilon$ is the perturbation budget. $\delta^{(m)}$ is initialized from a random seed specific to attacker $m$ and optimized over $E = 10$ epochs by maximizing the binary cross-entropy loss on the surrogate via Projected Gradient Descent (PGD) [44]. The optimization objective is:

$$\delta^{(m)} = arg \max_{\|\delta\|_\infty \leq \varepsilon} \frac{1}{|D^{(m)}|} \sum_{(x_i, y_i) \in D^{(m)}} \mathcal{L}_{BCE}(f_{surr}(x_i + \delta), y_i),$$

where $(x_i, y_i)$ refers to the training video and label pair in the dataset $D^{(m)}$ of attacker $m$.

**Supplementary Information**

Supplementary Information includes:

- Supplementary Figures S1-S10
- Methods on spatially multiplexed deepfake video detection using a diffractive decoder
- Cross-talk analysis of optical decoder architecture

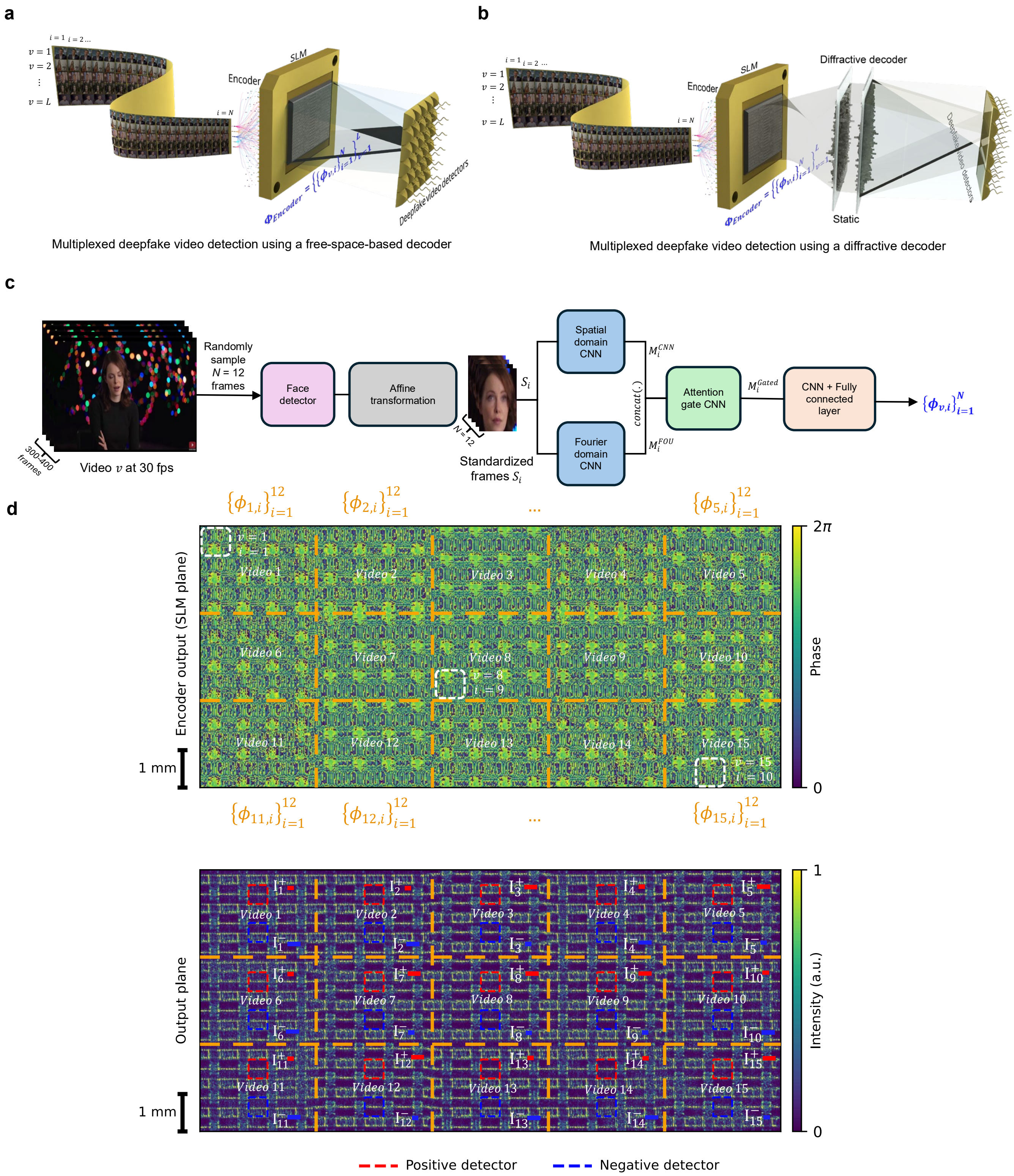

**Figure 1: Multiplexed deepfake video detection using a digital encoder and an optical decoder. a** Schematic of a multiplexed deepfake video detection with a digital encoder and a free-space-based static decoder. The incoming batch of $L$ video streams is first digitally encoded into a phase pattern $\mathbf{\Phi}$, which is then displayed on the SLM. After the encoded optical field at the SLM plane propagates through the free-space decoder, the intensity of the output field is recorded by a set of paired differential detectors (i.e., $2L$ output detectors in total). **b** Schematic of a multiplexed deepfake video detection with a digital encoder and a static diffractive decoder. The input optical field at the SLM plane propagates through free-space and is further modulated by $K$ phase-only diffractive layers before optical detection. **c** Digital encoding pipeline for phase generation. For each video $v$ in the batch, $N$ frames are randomly sampled and processed by a lightweight face detection network. The frames are then independently processed by the Fourier and spatial-domain CNN branches. The extracted features are concatenated and fused via an attention gate, and a final fully connected layer that maps the fused representation into a phase-pattern. **d** Snapshot of one batch of $L = 15$ videos, each with $N = 12$ frames. The top panel shows the phase-only encoder output displayed on the SLM. The bottom pattern shows the corresponding captured output intensity image. For each video $v$, paired positive (red, $I_v^+$) and negative (blue, $I_v^-$) detector regions are annotated. The mean pixel intensity value inside each detector is depicted by horizontal bars. Longer red bars ($I_v^+ > I_v^-$) indicate a fake video, while longer blue bars ($I_v^- > I_v^+$) indicate a real video. This image has been contrast-enhanced for better visualization using imadjust with lower and upper saturation limits of 0.05 and 0.95, respectively; however, raw image values are used for computing the intensity and classification scores throughout the manuscript.

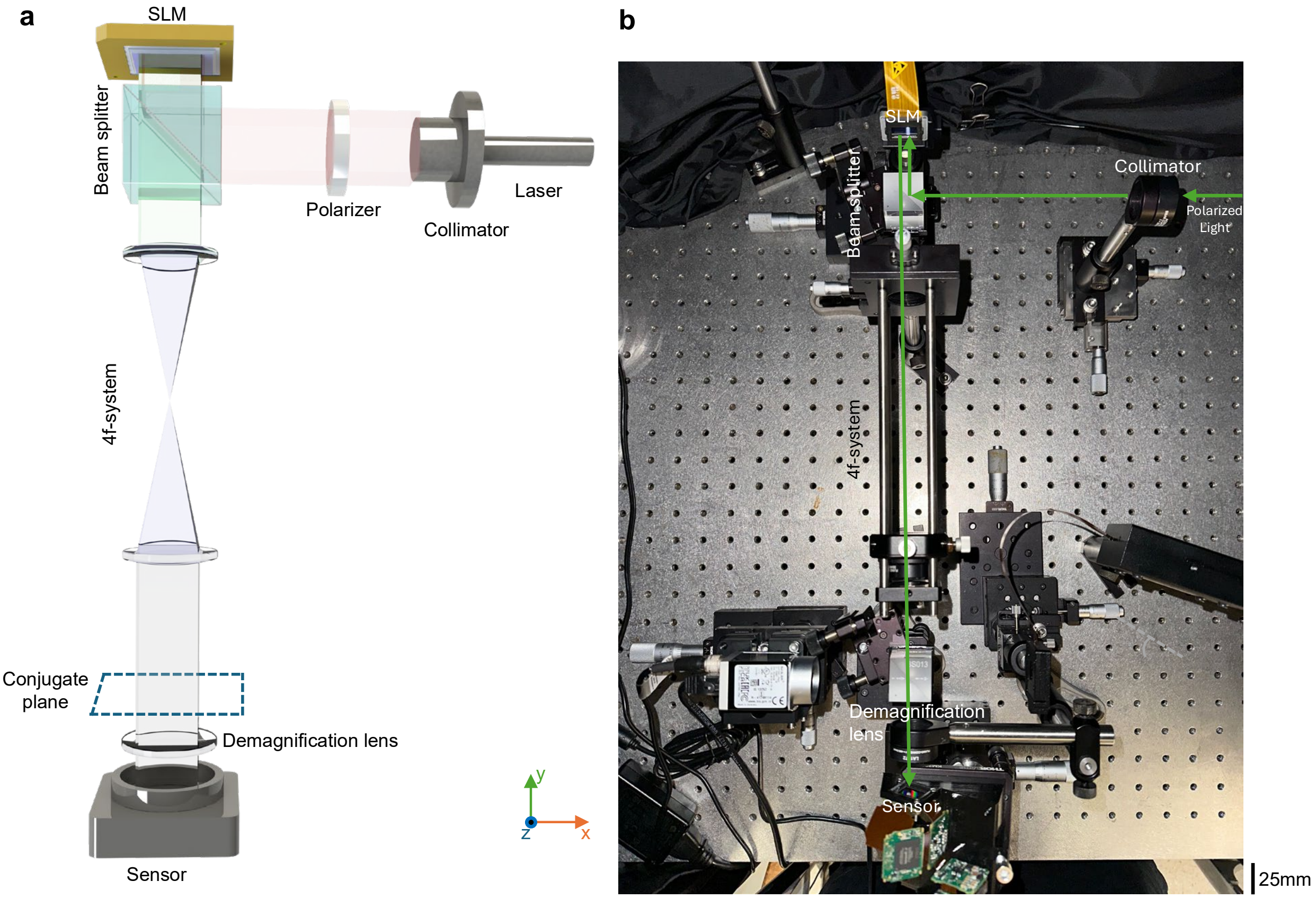

**Figure 2: Experimental demonstration of a multiplexed deepfake video detector. a** Schematic of the experimental set-up for multiplexed deepfake video detection with free-space-based decoder. The collimated laser light illuminates the SLM after being deflected by the beam splitter. Then a 4f-system is applied to conjugate the encoded wavefront to a position before the sensor plane. After the conjugate plane, a demagnification lens is used to capture the final detection intensity. **b** Photograph of the experimental set-up. The green arrow indicates the light path.

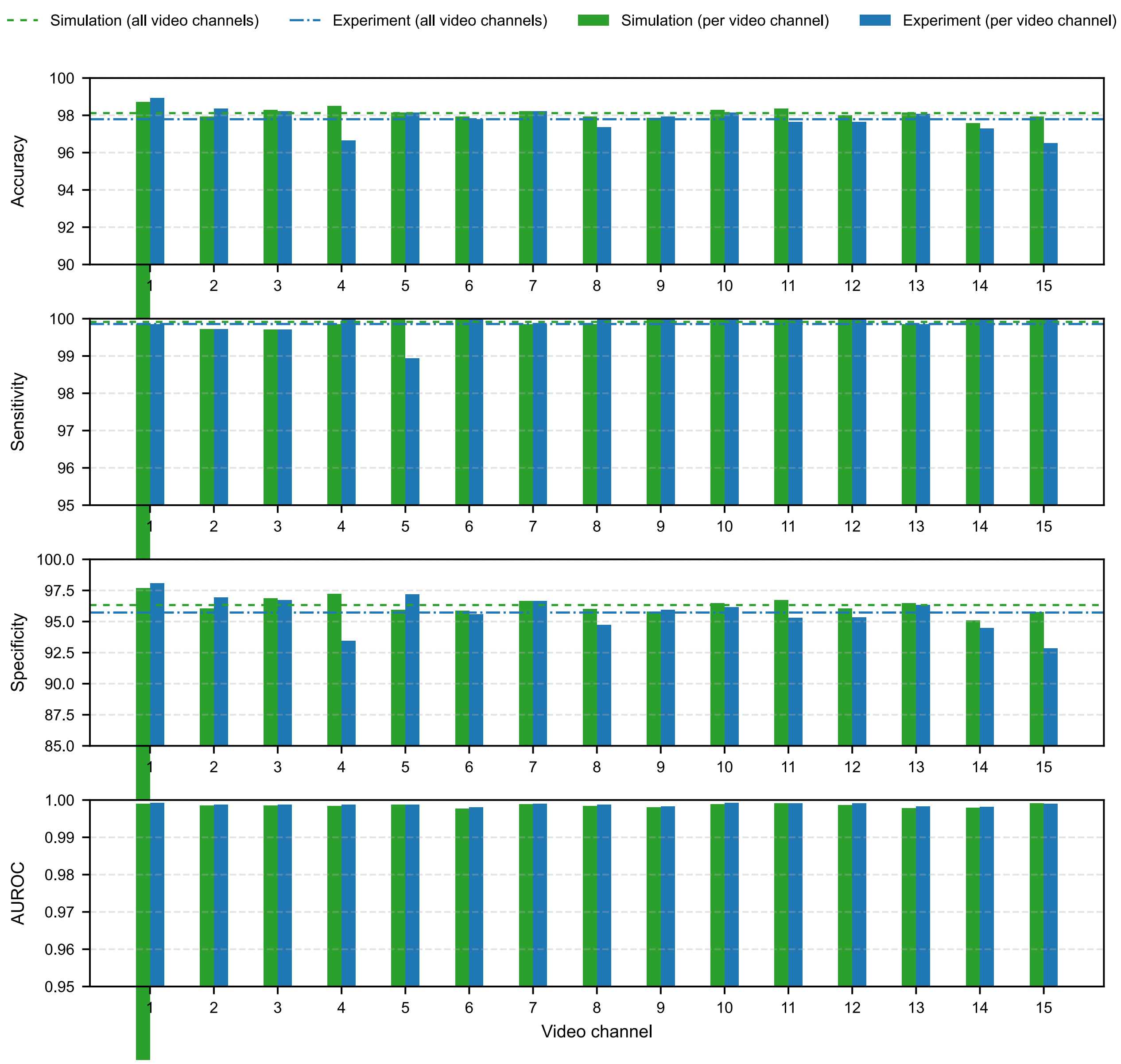

**Figure 3: Channel-wise performance of free-space-based multiplexed deepfake video detection model on the Celeb-DF dataset for $L = 15$ spatially-multiplexed design.** Accuracy, sensitivity, specificity and AUROC values of each of the $L = 15$ video channels are shown in the figure. The green and blue bars represent the simulation and experimental results of each video channel, respectively. The dashed lines indicate the overall performance across all channels. There is a strong agreement between the simulation and experimental results across all video channels. Overall accuracy is 98.12% ± 0.28% in simulations and 97.79% ± 0.62% in experiments, while specificity is 96.31% ± 0.63% in simulations and 95.72% ± 1.37% in experiments. Sensitivity remains at ~99.9% in both simulations and experimental results. Also see **Supplementary Fig. S2** for the $L = 18$ multiplexed design experimental and numerical results.

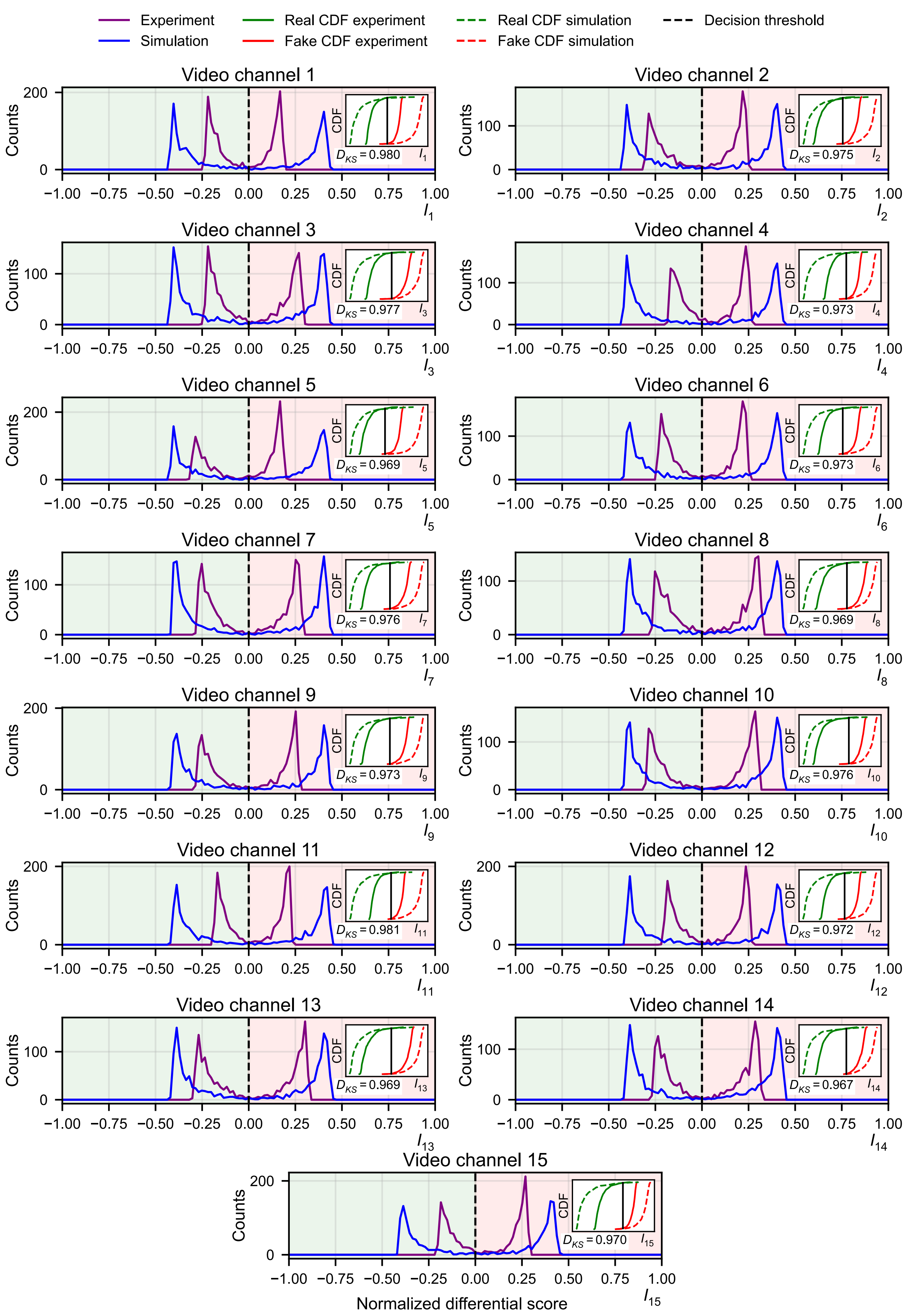

**Figure 4: Channel-wise normalized differential score distributions for real and fake**

**videos of the Celeb-DF dataset for $L = 15$ spatially multiplexed design**. Histogram of the normalized differential scores $I_v$ for video channels $v = 1, 2, \dots, 15$. Experimental (purple) and numerical (blue) results are shown. The decision threshold is indicated by the vertical dashed lines. Inset displays the empirical cumulative distribution functions (CDFs) of real (green) and fake (red) video scores in both experimental (solid) and simulation (dashed) results. For each channel, the Kolmogorov-Smirnov distance (vertical black bar) $D_{KS}$ between the real and fake score distributions in the corresponding experiment is reported below the inset, with a mean of 0.9733 and a standard deviation of 0.0042. Also see **Supplementary Fig. S3** for the $L = 18$ multiplexed design experimental and numerical results.

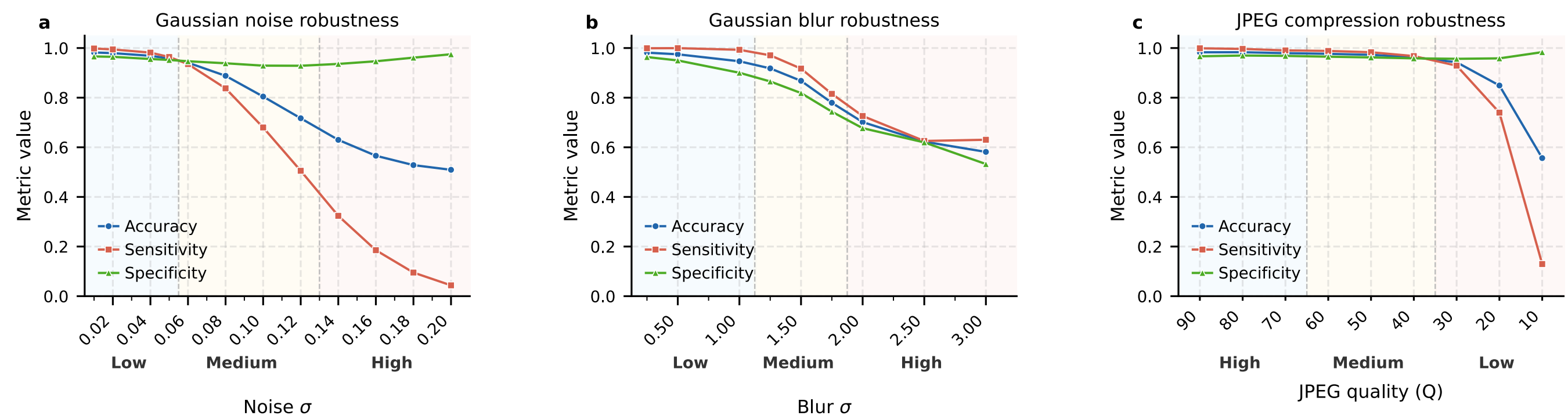

**Figure 5: Robustness of the $L = 15$ spatially multiplexed deepfake video detection model to frame-level degradations.** Detection performance is evaluated across three categories of image perturbations not encountered during training. Accuracy, sensitivity, and specificity are plotted as a function of **a** Gaussian noise standard deviation $\sigma$, **b** Gaussian blur standard deviation $\sigma$, and **c** JPEG compression quality factor $Q$ as defined by the JPEG standard. Shaded regions denote low (blue), medium (yellow), and high (red) perturbation regimes. **a** The system maintains sensitivity above 95% for $\sigma \leq 0.05$, after which it drops sharply under stronger noise levels, representing the most impactful degradation category. **b** Under Gaussian blur, accuracy remains above 94% and 78% in the mild and moderate regimes, respectively, degrading gradually to ~60% under strong blur. **c** All three metrics remain above 93% across most JPEG quality levels, with a notable drop only for $Q < 30$, where compression artifacts become visually apparent. Overall, the hybrid deepfake video processor maintains stable performance across mild and moderate degradation levels that are representative of practical real-world conditions.

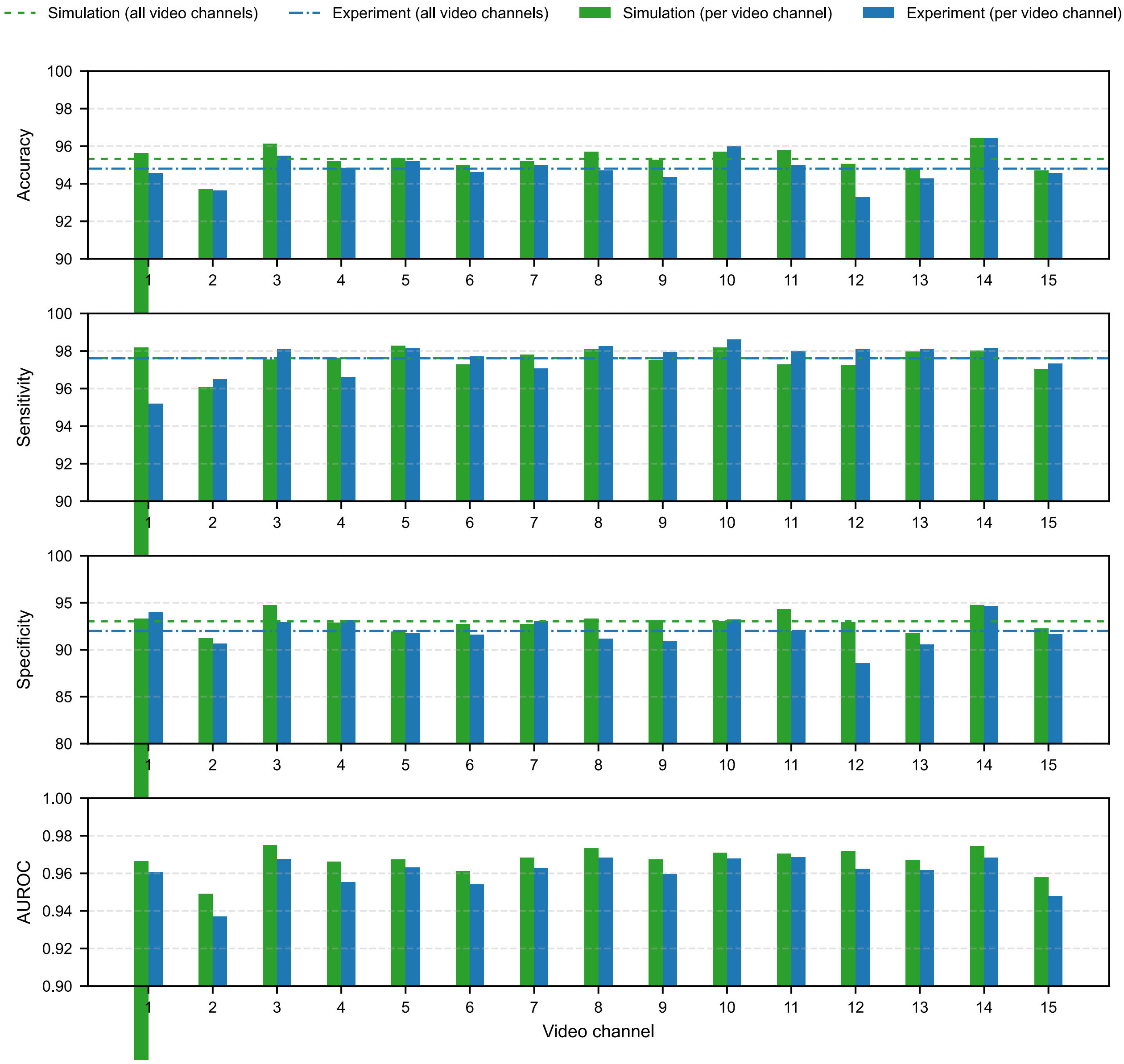

**Figure 6: Channel-wise transfer learning performance on VEO-3 (Gemini) generated videos for the $L = 15$ spatially multiplexed design.** Accuracy, sensitivity, specificity and AUROC values of each of the $L = 15$ video channels are shown in the figure. The green and blue bars represent simulation and experimental results of each channel, respectively. The dashed lines indicate the overall performance across all video channels. Overall accuracy is 95.32% ± 0.63% in simulations and 94.80% ± 0.78% in experiments, while specificity is 93.02% ± 0.99% in simulations and 92.00% ± 1.50% in experiments. Sensitivity remains above ~97.5% across all channels in both simulations and experiments. Also see **Supplementary Fig. S7** for the $L = 18$ design experimental and numerical results.

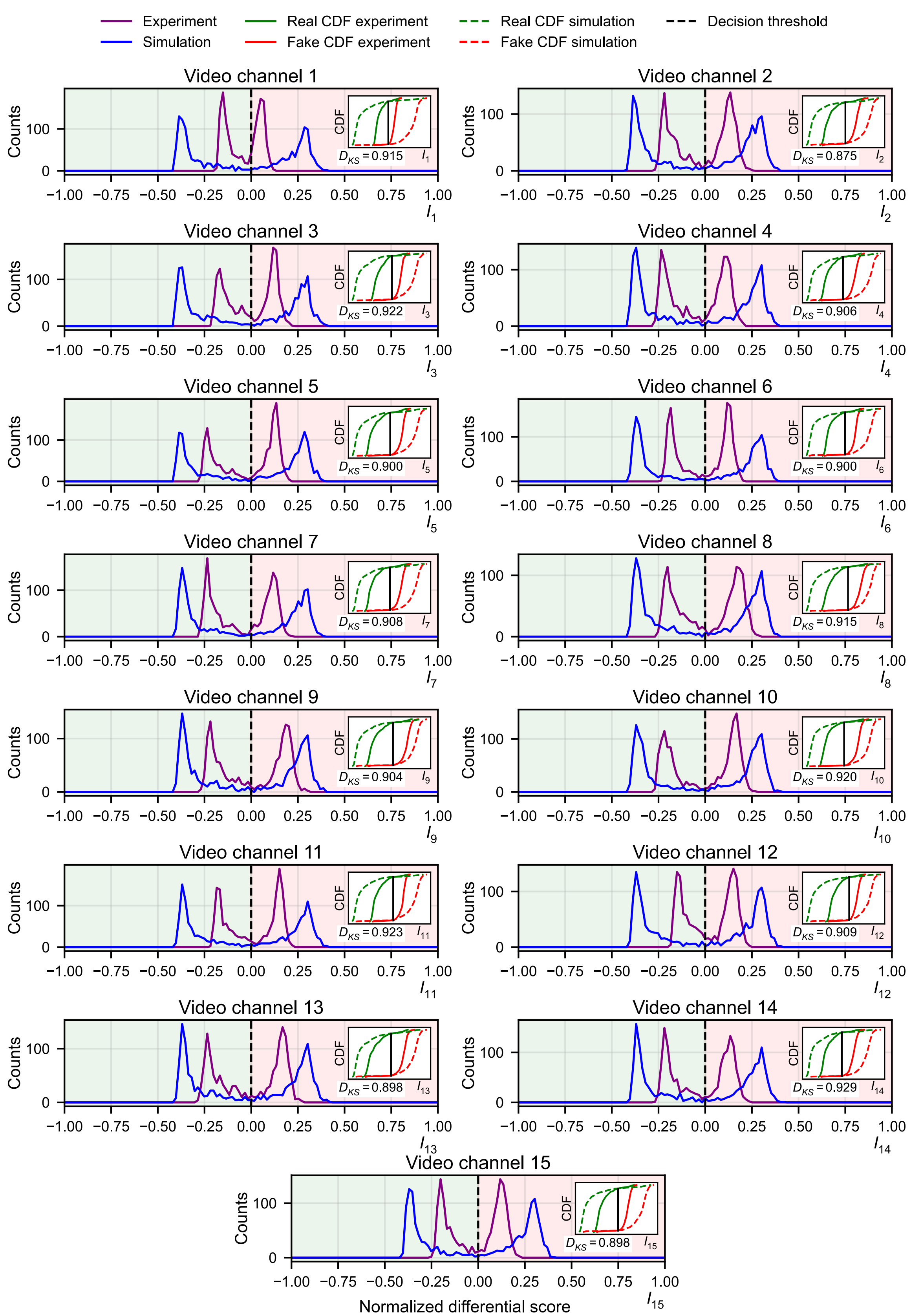

**Figure 7: Channel-wise normalized differential score distributions for VEO-3 (Gemini)**

**generated videos for the $L = 15$ multiplexed design**. Histogram of the normalized differential scores $I_v$ for video channels $v = 1, 2, \ldots, 15$. Experimental (purple) and numerical (blue) results are shown. The decision threshold is indicated by the vertical dashed lines. Inset displays the empirical CDFs of real (green) and fake (red) video scores in both experimental (solid) and simulation (dashed) results. For each channel, the Kolmogorov-Smirnov distance (vertical black bar) $D_{KS}$ between the real and fake score distributions in the corresponding experiment is reported below the inset, with a mean of 0.9087 and a standard deviation of 0.0138. Also see **Supplementary Fig. S8** for the $L = 18$ multiplexed design experimental and numerical results.

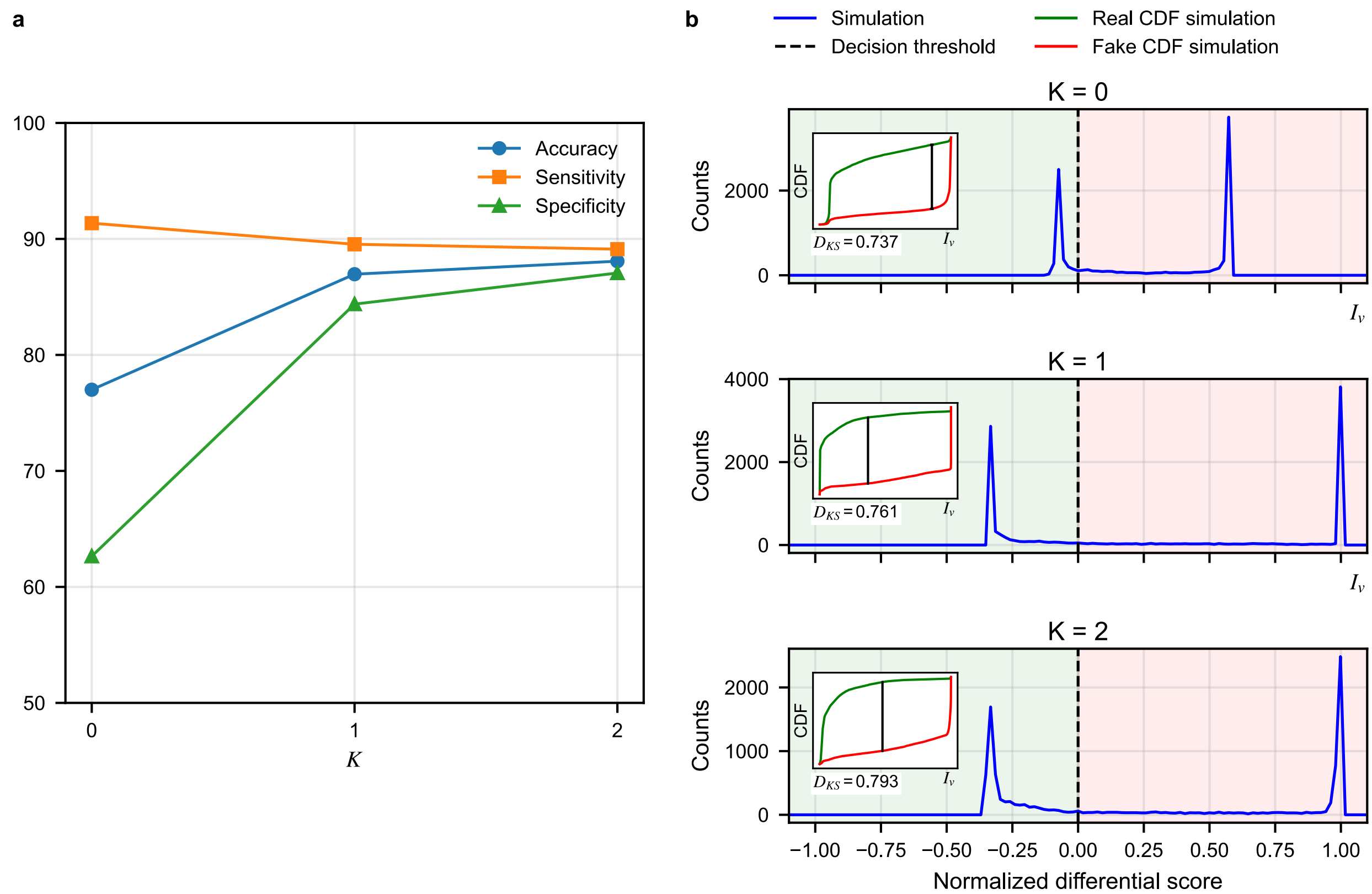

**Figure 8**: **Simulation results on the DeepSpeak dataset across $K$ = 0, 1 and 2. a** Accuracy, sensitivity, and specificity as a function of diffractive decoder depth (*i.e.* $K$). Adding $K$ = 2 static diffractive layers that are structurally optimized improves specificity and accuracy by ~24% and ~11%, respectively, with a ~2% drop in sensitivity compared to the vanilla decoder with $K = 0$. **b** Normalized differential scores are computed for each diffractive depth $K$. For each $K$, the Kolmogorov-Smirnov distance $D_{KS}$ between the real and fake score distributions is reported below the inset. Increasing the number of optimized diffractive layers leads to a higher $D_{ks}$ indicating better statistical separation between real and fake videos and more confident classification performance.